\documentclass[twocolumn]{article} 

\usepackage{amsmath, amsthm, amssymb, amsfonts}

\usepackage[numbers,square]{natbib}
\usepackage[utf8]{inputenc}	
\usepackage[T1]{fontenc}	
\usepackage{xcolor}		
\usepackage[colorlinks = true,
            linkcolor = purple,
            urlcolor  = blue,
            citecolor = cyan,
            anchorcolor = black]{hyperref}	
\usepackage{booktabs} 		
\usepackage{nicefrac}		
\usepackage{microtype}		
\usepackage{lineno}		
\usepackage{float}			

\usepackage{lipsum}		

\usepackage{newfloat}
\DeclareFloatingEnvironment[name={Supplementary Figure}]{suppfigure}
\usepackage{sidecap}
\sidecaptionvpos{figure}{c}

\usepackage{titlesec}
\titlespacing\section{0pt}{12pt plus 3pt minus 3pt}{1pt plus 1pt minus 1pt}
\titlespacing\subsection{0pt}{10pt plus 3pt minus 3pt}{1pt plus 1pt minus 1pt}
\titlespacing\subsubsection{0pt}{8pt plus 3pt minus 3pt}{1pt plus 1pt minus 1pt}

\usepackage{tikz,xcolor,hyperref}

\definecolor{lime}{HTML}{A6CE39}
\DeclareRobustCommand{\orcidicon}{
	\begin{tikzpicture}
	\draw[lime, fill=lime] (0,0) 
	circle [radius=0.16] 
	node[white] {{\fontfamily{qag}\selectfont \tiny ID}};
	\draw[white, fill=white] (-0.0625,0.095) 
	circle [radius=0.007];
	\end{tikzpicture}
	\hspace{-2mm}
}
\foreach \x in {A, ..., Z}{\expandafter\xdef\csname orcid\x\endcsname{\noexpand\href{https://orcid.org/\csname orcidauthor\x\endcsname}
			{\noexpand\orcidicon}}
}

\title{LACTOSE: Linear Array of Conditions, TOpologies with Separated Error-backpropagation - The Differentiable "IF" Conditional for Differentiable Digital Signal Processing}

\usepackage{authblk}

\author[1\thanks{\tt{chris.clarke@uec.ac.jp}}]{Christopher Johann Clarke}

\affil[1]{The University of Electro-Communications	}

\usepackage{algorithm}
\usepackage{algorithmic}

\begin{document}

\maketitle

\begin{abstract}
There has been difficulty utilising conditional statements as part of the neural network graph (e.g. if input $> x$, pass input to network $N$). This is due to the inability to backpropagate through branching conditions. The Linear Array of Conditions, TOpologies with Separated Error-backpropagation (LACTOSE) Algorithm addresses this issue and allows the conditional use of available machine learning layers for supervised learning models. In this paper, the LACTOSE algorithm is applied to a simple use of DDSP, however, the main point is the development of the "if" conditional for DDSP use. The LACTOSE algorithm stores trained parameters for each user-specified numerical range and loads the parameters dynamically during prediction.
\end{abstract}
\vspace{0.35cm}



\section{Introduction}\label{sec:introduction}
The utilization of conditional ``if" functions in Differential Digital Signal Processing (DDSP) presents unique difficulties that must be overcome to achieve accurate and efficient results. This paper focuses on the challenge of allowing training to take place in the presence of these functions. The authors present the LACTOSE Algorithm, which allows the use of differentiable "if" conditions to address these difficulties. The paper will explore the potential of the LACTOSE Algorithm to improve DDSP and highlight areas where further research is needed.

\begin{figure}[t]
    \centering
    \includegraphics[width=\columnwidth]{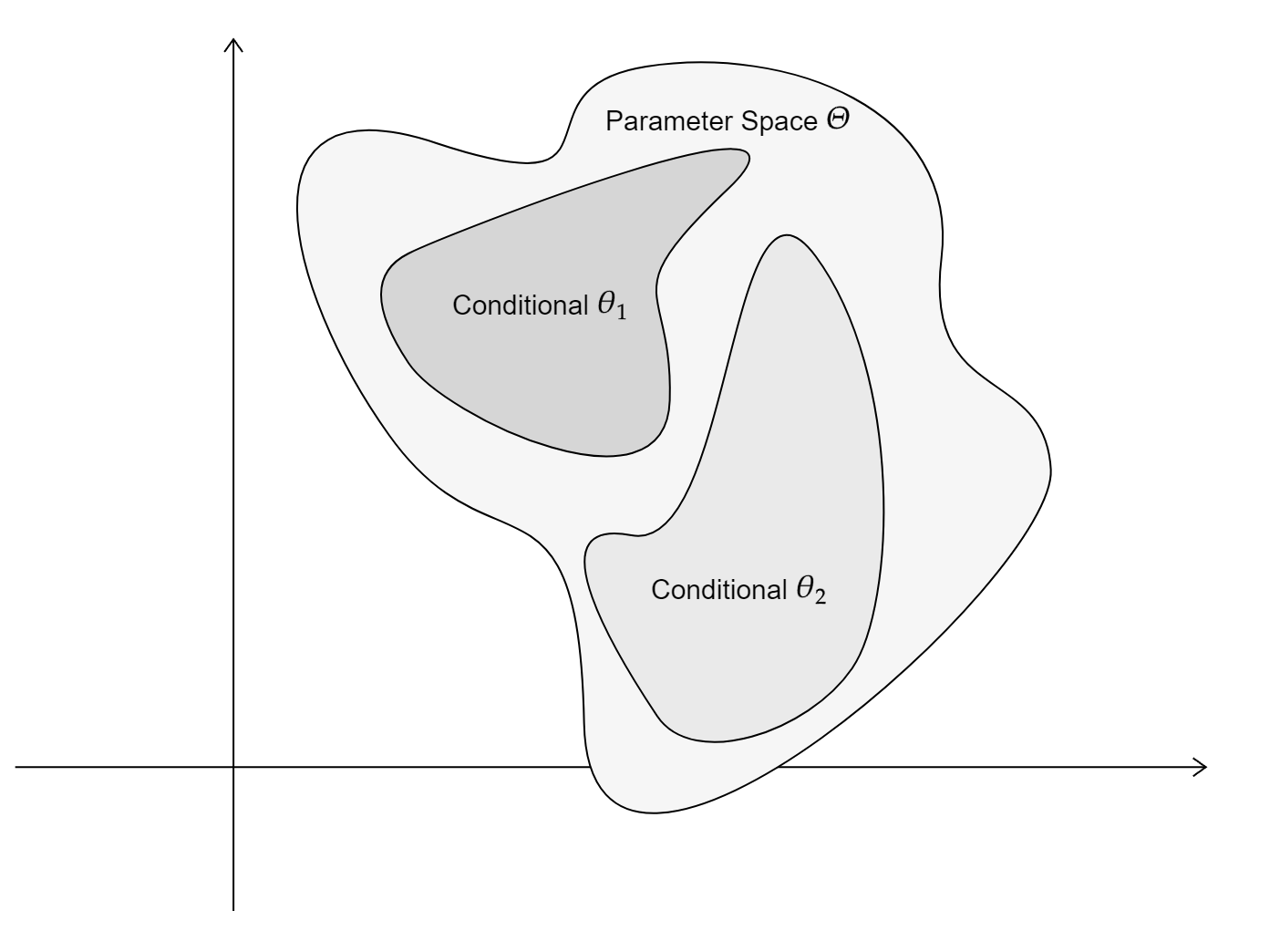}
    \caption{A visualisation of a dimension-reduced model parameter space. When passing the conditions as an input to the model, or if the model has to learn the condition parameter, the model has to train for a parameter space that encompasses all conditions. }
    \label{fig:combinedparams}
\end{figure}

Unless one is using machine learning methods, such as Decision Trees or Markov Modelling, that do not require error-backpropagation~\cite{10.1007/3-540-70659-3_2} \textemdash it is an issue that branching condition statements are not differentiable, and are not used in the model architecture. Conditional Modelling (CM) has been investigated in a variety of different manners. In the case of Conditional Random Fields (CRF), it is usually attributed to a likelihood parameter. The CRF was proposed to as a solution to the limitations of Hidden Markov Models and Maximum Entropy Markov Models~\cite{8167121}. CRFs are a construction of a graphical model for which each prediction can be contextually inferred based on neighbouring samples~\cite{8598722, MAL-013}. This has been extended by adding a trainable hidden parameterised gate layer in the middle to form Conditional Neural Fields~\cite{NIPS2009_e820a45f}.

\begin{figure}[t]
    \centering
    \includegraphics[width=\columnwidth]{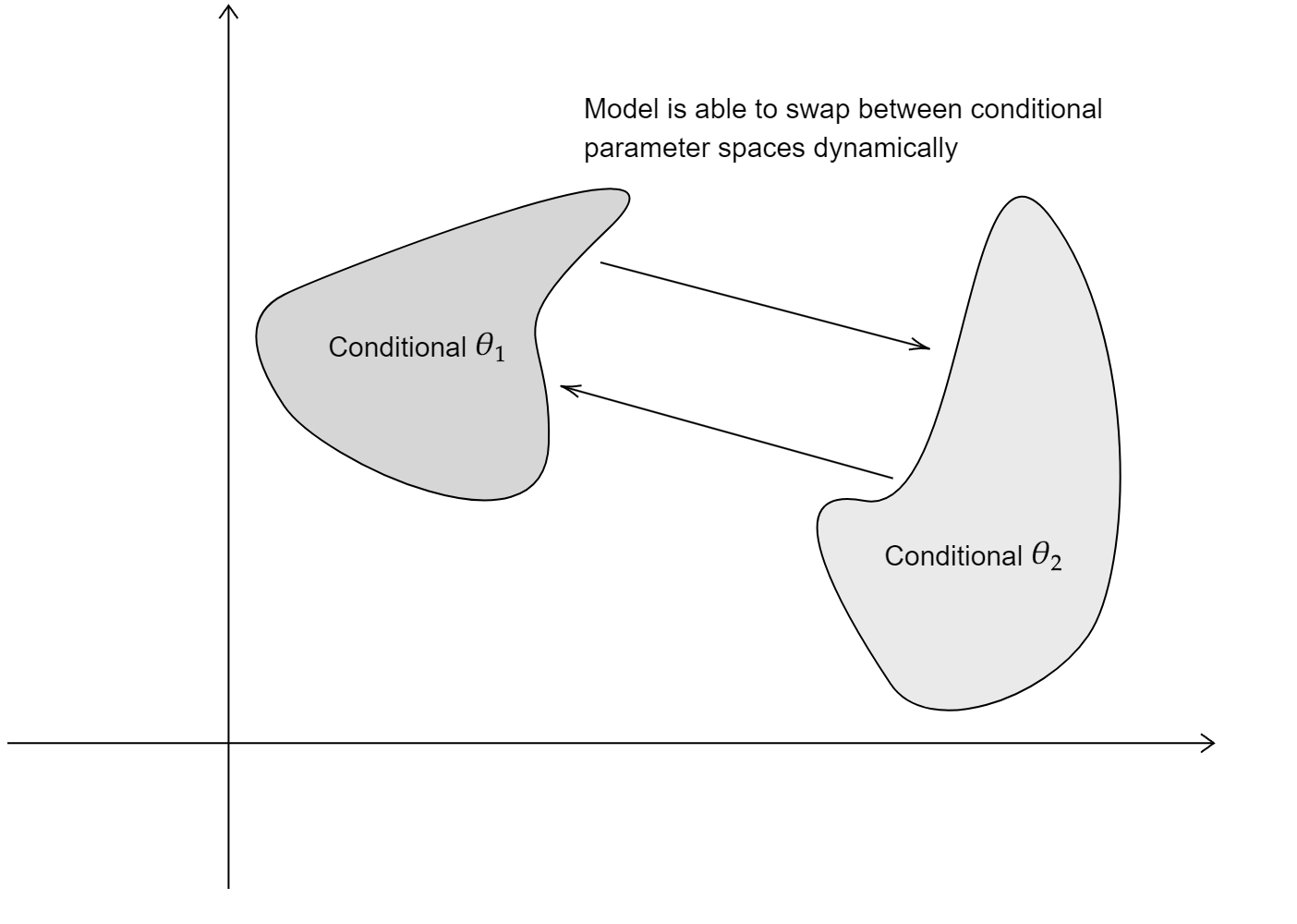}
    \caption{This visualisation of the dimension-reduced parameter space shows the separated parameter spaces that pertain to each branching condition. The LACTOSE algorithm allows for the model to dynamically swap between each parameter space, and thus not requiring the model to train for an encompassing parameter space.}
    \label{fig:separatedparams}
\end{figure}

\begin{figure*}[t!]
    \centering
    \includegraphics[width=\linewidth]{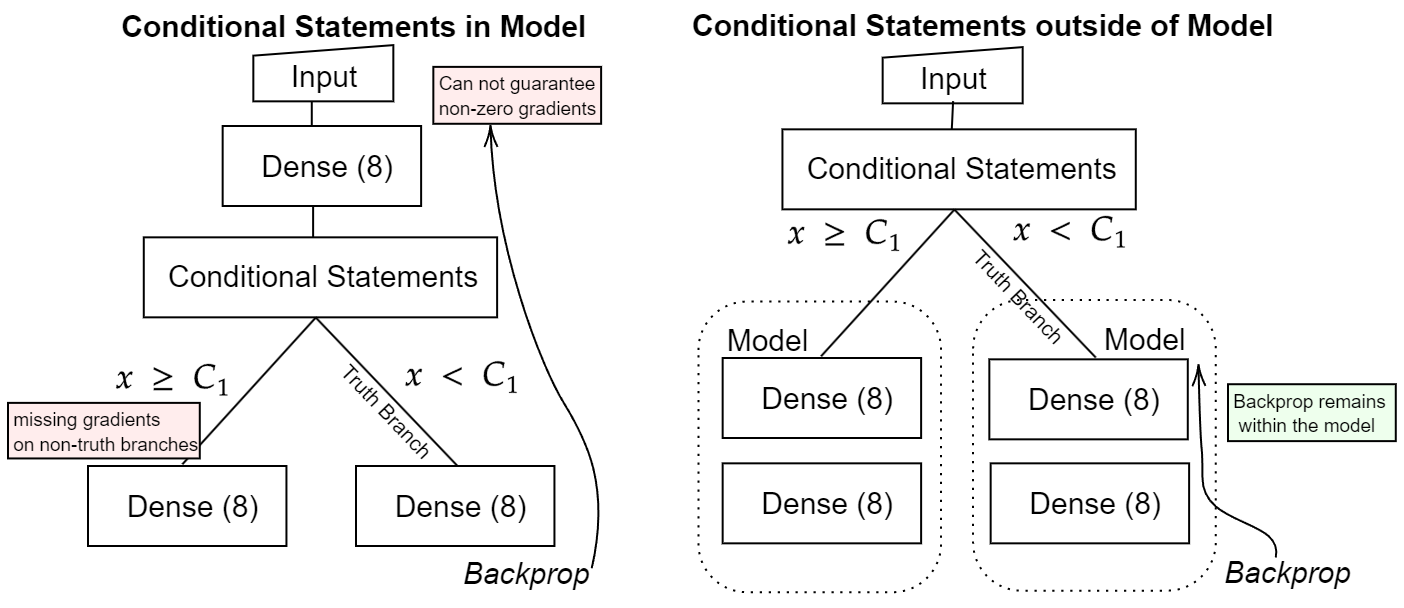}
    \caption{The figure on the left demonstrates the issues faced by branching condition statements inside a model. The figure on the right demonstrates the proposed approach that LACTOSE is designed with.}
    \label{fig:condition}
\end{figure*}

\begin{algorithm*}[t]
\caption{LACTOSE Algorithm.}
\label{alg:LACTOSE}
\textbf{Input}: $x$, $y$\\
\textbf{Parameters}: Model Parameters $\theta_1$,...,$\theta_N$,\\ Conditions $C_1$,...,$C_N$  \\
\textbf{Output}: $\hat{y}$
\begin{algorithmic}[1] 
\STATE Model Input = $x$.
\STATE Model Truth = $y$
\IF {$x = C_N$}
    \RETURN {$\theta_N$}
\ENDIF
\STATE Model $\gets \theta_N$
\STATE Prediction $\hat{y} \gets$ Model$(x)$
\STATE Loss Function $\gets$ $(y$,$\hat{y})$
\STATE \textbf{return} Loss
\STATE Model $\gets$ Optimizer(Loss)
\STATE \textbf{save} Model Parameters $\gets$ new $\theta_N$
\end{algorithmic}
\end{algorithm*}

Other work has developed Conditional Neural Processes (CNP), which are an extension of Gaussian Processes. CNPs seek to parameterise conditional processes with respect to a prior process. In doing so, CNPs are extensible in their functional flexibility and scalability, as their inner process can be computed in $\mathcal{O}(1)$~\cite{garnelo2018conditional}. CNPs have shown to perform comparatively (if not better than) Gaussian Processes and other Bayesian optimization methods~\cite{9186363}.

There is also the possibility to avoid the issue of error-backpropagation by allowing the model to learn the conditions within the state space. Constrained Conditional Models (CCM) have a trainable offset penalty~\cite{chang2012structured}, a trainable Action (one-hot)~\cite{DBLP:journals/corr/abs-1907-01749}, or a learnt parameter vector~\cite{DBLP:journals/corr/abs-1809-06127}.

\begin{figure*}[t]
    \centering
    \includegraphics[width=\linewidth]{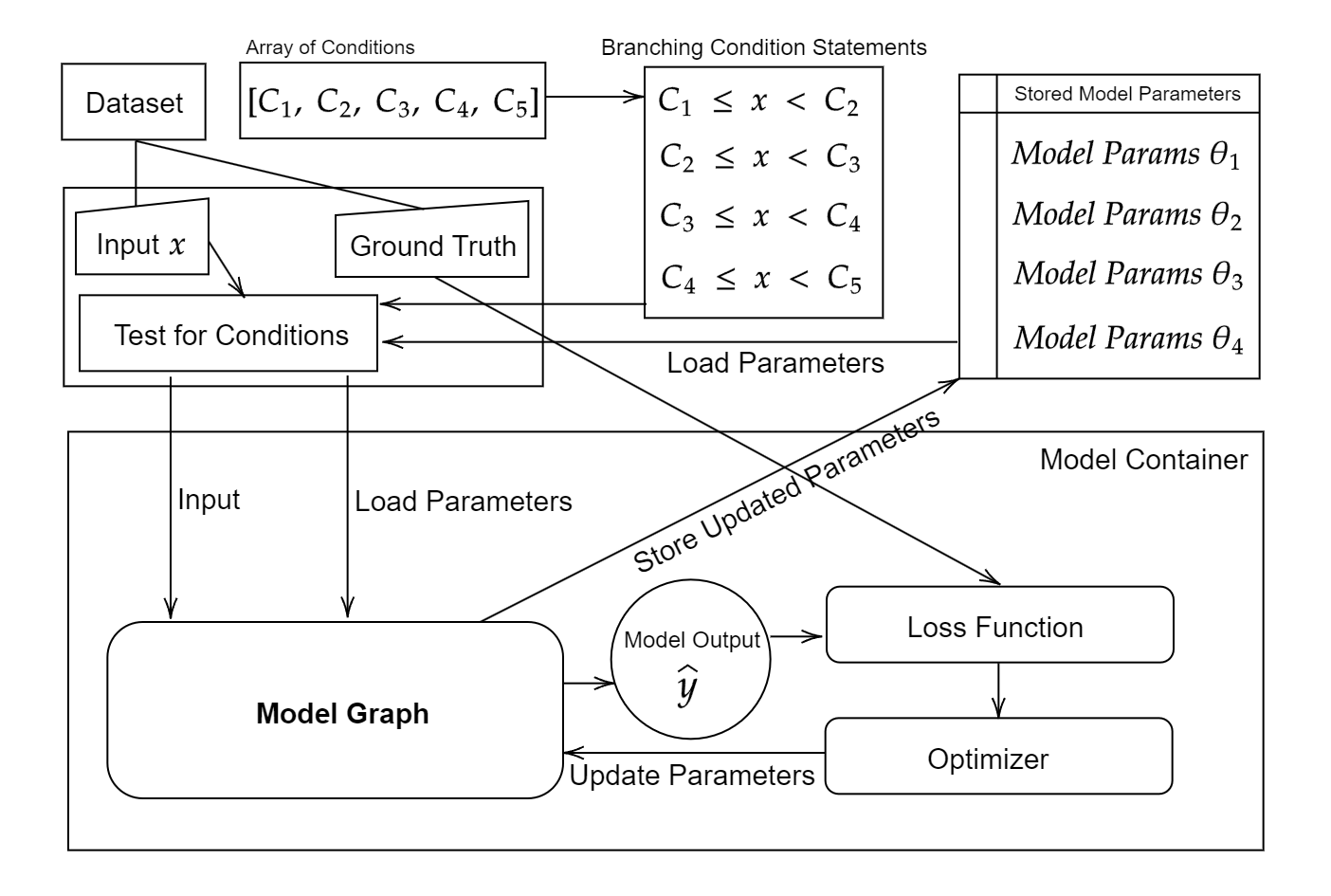}
    \caption{Procedures behind the LACTOSE algorithm. The box in red represents the static Tensorflow graph. Conditions are hosted externally from the graph and can therefore make use of symbolic inputs.}
    \label{fig: LACTOSE}
\end{figure*}

Conditional Variational Autoencoders and Conditional Seq 2 Seq Frameworks are another method that makes use of conditions. These methods either include conditions to the model as part of the input~\cite{makris2021generating, DBLP:journals/corr/ZhaoZE17}, a side input to the encoder~\cite{https://doi.org/10.48550/arxiv.1911.13220, nguyen2021conditional}, or an input to the decoder~\cite{8950195}.

In the methods mentioned, the conditional statements are either provided to the model as an input (concatenated or side-input) or learnt by the network (as in CCM). In both of these situations, the trained parameter space will need to encompass all of the existing conditions, as shown by Figure~\ref{fig:combinedparams}. This means that a model will need an increasingly greater parameter space if the parameter spaces corresponding to each condition are ``spatially'' further apart. This greater parameter space is usually achieved by increasing the number of parameters in the model, provided that an encompassing parameter space exists. However, as alluded to earlier, a larger model architecture will increase inference time. A solution to this, as shown by Figure~\ref{fig:separatedparams}, is to have the model dynamically swap between the parameter space associated to each condition, thereby reducing the size of necessary parameter space. This approach requires the model creation to receive an immutable set of branching condition statements, this can be informed either a priori, through empirical deduction from analyses, or domain knowledge and intuition.

In this paper we will confront the issue associated with an encompassing and increased parameter space by proposing the Linear Array of Conditions, TOpologies with Separated Error-backpropagation (LACTOSE) Algorithm in the next section. Finally, a conclusion is offered. 

\section{LACTOSE Algorithm} 
\label{lactose}
The LACTOSE Algorithm addresses the issues faced when applying branching condition statements. Consider the two cases in Figure~\ref{fig:condition}, which demonstrate the issues faced by branching condition statements inside a model (graph):

Without assumptions or knowledge of the automatic differentiation framework and dataflow of the machine learning system used, describing the graph on the left in Figure~\ref{fig:condition}: To give an overview of how the data might be passed down the network, an input is passed to the first Dense (Fully Connected) layer. The output of this Dense layer is tested against the conditional statements and a truth branch is chosen from the available paths, where the data continues down the network. 

Firstly, a problem arises during error-backpropagation when the automatic differentiation framework can not guarantee a non-zero gradient (or produces a NaN valued gradient) when differentiating past a branching condition statement. 

Secondly, because of the possibility of zero-valued gradients, the first layer (right after the input) will not have any gradients to adjust its parameters. 

Lastly, there will be missing gradients for the other models continuing from the non-truth branches, as the error-backpropagation framework has no access to these models.

In the graph of the model, any variable that is not ``compiled'' with a fixed value is known as a symbolic variable~\cite{WhatareS44:online}. When an input is passed from the dataset into the model graph, the symbolic variable is assigned this input and the model graph will act on that variable. When branching condition statements are used in the model graph, a symbolic variable is used. This symbolic variable would be used to test against the condition, and the consequent or alternatives would be returned. The automatic differentiation framework cannot guarantee a non-zero derivative if the branching condition statement depends on the value associated with a symbolic variable. Furthermore, the automatic differentiation framework can only act upon the executed branch (truth branch) path. To prevent this, most automatic differentiation frameworks require the computational graph to be fixed.

The LACTOSE algorithm directly addresses the issues associated with backpropagation through branching condition statements. The library was implemented in Tensorflow~\cite{tensorflow2015-whitepaper}. Figure~\ref{fig: LACTOSE} presents the computational procedures behind LACTOSE.

The algorithm takes a dataset and an array of conditions \textemdash represented by points on the number line\textemdash as input. Upon initialisation, the model parameters are stored. The number of copies of model parameters depend on the number of conditions. This can be depicted two ways. Figure~\ref{fig:condition} illustrates this as separate models for each branch of the conditions, while Figure~\ref{fig: LACTOSE} shows that in practice, these copies of the model paramaters are stored outside of the model graph. LACTOSE hosts the conditions and stored model parameters outside of the static ``compiled'' Tensorflow graph. The truth branch is then derived from the conditions, and the respective model parameters associated with that truth branch are dynamically loaded into the model before running a prediction. The loss is then calculated and the error is propagated within the graph. Lastly, the new model parameters for this truth branch are updated in the stored model parameters list.  
Formally it is as written in Algorithm~\ref{alg:LACTOSE}.

\section{Conclusion}
An algorithm for using branching condition statements has been developed and implemented as a library using the Tensorflow framework. In this paper, a survey of Condition Modelling algorithms and architectures was done, showing the various other efforts and solutions that have been put forth to integrate conditional statements, conditions, and learnt conditions into neural network model architectures. The LACTOSE algorithm was described and a preliminary methodology of how to approach problems was demonstrated. 

However, the algorithm is currently only able to train and inference at a rate of a batch size of 1, as the model needs to retrieve parameters and store parameters per training loop. Future work will be put into this algorithm to allow it to train with a larger batch size.

More work will also be put into this model such that it will be able to mask certain layers during training and inference. For example, a situation might arise such that the extremities of the branching conditions require a CNN, but the center regions require an LSTM. This can be done by setting a dropout on each layer, and masking of individual layers as the situation necessitates \textemdash allowing for even greater granularity and modularity in the side-effects of the branching condition statements. In addition, this method of channelling could potentially be exploited as an embedded feature extraction mechanism for 1D inputs. The different parameter spaces are associated with different input ranges.

The high-level extensible nature of the interface allows for the development of policies to perform search for the optimal branching condition statements given a certain dataset. 



\normalsize
\bibliography{references}

\begin{thebibliography}{17}
\providecommand{\natexlab}[1]{#1}
\providecommand{\url}[1]{\texttt{#1}}
\expandafter\ifx\csname urlstyle\endcsname\relax
  \providecommand{\doi}[1]{doi: #1}\else
  \providecommand{\doi}{doi: \begingroup \urlstyle{rm}\Url}\fi

\bibitem[Dietterich(2002)]{10.1007/3-540-70659-3_2}
Thomas~G. Dietterich.
\newblock Machine learning for sequential data: A review.
\newblock In Terry Caelli, Adnan Amin, Robert P.~W. Duin, Dick de~Ridder, and Mohamed Kamel, editors, \emph{Structural, Syntactic, and Statistical Pattern Recognition}, pages 15--30, Berlin, Heidelberg, 2002. Springer Berlin Heidelberg.
\newblock ISBN 978-3-540-70659-5.

\bibitem[Liliana and Basaruddin(2017)]{8167121}
Dewi~Yanti Liliana and Chan Basaruddin.
\newblock A review on conditional random fields as a sequential classifier in machine learning.
\newblock In \emph{2017 International Conference on Electrical Engineering and Computer Science (ICECOS)}, pages 143--148, 2017.
\newblock \doi{10.1109/ICECOS.2017.8167121}.

\bibitem[Wang et~al.(2019)Wang, Yuan, and Liu]{8598722}
Qiurui Wang, Chun Yuan, and Yan Liu.
\newblock Learning deep conditional neural network for image segmentation.
\newblock \emph{IEEE Transactions on Multimedia}, 21\penalty0 (7):\penalty0 1839--1852, 2019.
\newblock \doi{10.1109/TMM.2018.2890360}.

\bibitem[Sutton and McCallum(2012)]{MAL-013}
Charles Sutton and Andrew McCallum.
\newblock An introduction to conditional random fields.
\newblock \emph{Foundations and Trends® in Machine Learning}, 4\penalty0 (4):\penalty0 267--373, 2012.
\newblock ISSN 1935-8237.
\newblock \doi{10.1561/2200000013}.
\newblock URL \url{http://dx.doi.org/10.1561/2200000013}.

\bibitem[Peng et~al.(2009)Peng, Bo, and Xu]{NIPS2009_e820a45f}
Jian Peng, Liefeng Bo, and Jinbo Xu.
\newblock Conditional neural fields.
\newblock In Y.~Bengio, D.~Schuurmans, J.~Lafferty, C.~Williams, and A.~Culotta, editors, \emph{Advances in Neural Information Processing Systems}, volume~22. Curran Associates, Inc., 2009.
\newblock URL \url{https://proceedings.neurips.cc/paper/2009/file/e820a45f1dfc7b95282d10b6087e11c0-Paper.pdf}.

\bibitem[Garnelo et~al.(2018)Garnelo, Rosenbaum, Maddison, Ramalho, Saxton, Shanahan, Teh, Rezende, and Eslami]{garnelo2018conditional}
Marta Garnelo, Dan Rosenbaum, Christopher Maddison, Tiago Ramalho, David Saxton, Murray Shanahan, Yee~Whye Teh, Danilo Rezende, and SM~Ali Eslami.
\newblock Conditional neural processes.
\newblock In \emph{International Conference on Machine Learning}, pages 1704--1713. PMLR, 2018.

\bibitem[Luo et~al.(2022)Luo, Chen, Li, and Zhang]{9186363}
Jianping Luo, Liang Chen, Xia Li, and Qingfu Zhang.
\newblock Novel multitask conditional neural-network surrogate models for expensive optimization.
\newblock \emph{IEEE Transactions on Cybernetics}, 52\penalty0 (5):\penalty0 3984--3997, 2022.
\newblock \doi{10.1109/TCYB.2020.3014126}.

\bibitem[Chang et~al.(2012)Chang, Ratinov, and Roth]{chang2012structured}
Ming-Wei Chang, Lev Ratinov, and Dan Roth.
\newblock Structured learning with constrained conditional models.
\newblock \emph{Machine learning}, 88\penalty0 (3):\penalty0 399--431, 2012.

\bibitem[Cai et~al.(2019)Cai, Yang, Zhang, Qin, and Li]{DBLP:journals/corr/abs-1907-01749}
Zexin Cai, Yaogen Yang, Chuxiong Zhang, Xiaoyi Qin, and Ming Li.
\newblock Polyphone disambiguation for mandarin chinese using conditional neural network with multi-level embedding features.
\newblock \emph{CoRR}, abs/1907.01749, 2019.
\newblock URL \url{http://arxiv.org/abs/1907.01749}.

\bibitem[Makris et~al.(2018)Makris, Kaliakatsos{-}Papakostas, and Kermanidis]{DBLP:journals/corr/abs-1809-06127}
Dimos Makris, Maximos~A. Kaliakatsos{-}Papakostas, and Katia~Lida Kermanidis.
\newblock Deepdrum: An adaptive conditional neural network.
\newblock \emph{CoRR}, abs/1809.06127, 2018.
\newblock URL \url{http://arxiv.org/abs/1809.06127}.

\bibitem[Makris et~al.(2021)Makris, Agres, and Herremans]{makris2021generating}
Dimos Makris, Kat~R Agres, and Dorien Herremans.
\newblock Generating lead sheets with affect: A novel conditional seq2seq framework.
\newblock In \emph{2021 International Joint Conference on Neural Networks (IJCNN)}, pages 1--8. IEEE, 2021.

\bibitem[Zhao et~al.(2017)Zhao, Zhao, and Esk{\'{e}}nazi]{DBLP:journals/corr/ZhaoZE17}
Tiancheng Zhao, Ran Zhao, and Maxine Esk{\'{e}}nazi.
\newblock Learning discourse-level diversity for neural dialog models using conditional variational autoencoders.
\newblock \emph{CoRR}, abs/1703.10960, 2017.
\newblock URL \url{http://arxiv.org/abs/1703.10960}.

\bibitem[Montserrat et~al.(2019)Montserrat, Bustamante, and Ioannidis]{https://doi.org/10.48550/arxiv.1911.13220}
Daniel~Mas Montserrat, Carlos Bustamante, and Alexander Ioannidis.
\newblock Class-conditional vae-gan for local-ancestry simulation, 2019.
\newblock URL \url{https://arxiv.org/abs/1911.13220}.

\bibitem[Nguyen et~al.(2021)Nguyen, Nguyen, Joty, and Li]{nguyen2021conditional}
Thanh-Tung Nguyen, Xuan-Phi Nguyen, Shafiq Joty, and Xiaoli Li.
\newblock A conditional splitting framework for efficient constituency parsing.
\newblock \emph{arXiv preprint arXiv:2106.15760}, 2021.

\bibitem[Zhu et~al.(2021)Zhu, Peng, and Wang]{8950195}
Jinlin Zhu, Guohao Peng, and Danwei Wang.
\newblock Dual-domain-based adversarial defense with conditional vae and bayesian network.
\newblock \emph{IEEE Transactions on Industrial Informatics}, 17\penalty0 (1):\penalty0 596--605, 2021.
\newblock \doi{10.1109/TII.2020.2964154}.

\bibitem[Gordon(2019)]{WhatareS44:online}
Josh Gordon.
\newblock What are symbolic and imperative apis in tensorflow 2.0? — the tensorflow blog.
\newblock \url{https://blog.tensorflow.org/2019/01/what-are-symbolic-and-imperative-apis.html}, January 2019.
\newblock (Accessed on 08/13/2022).

\bibitem[Abadi et~al.(2015)Abadi, Agarwal, Barham, Brevdo, Chen, Citro, Corrado, Davis, Dean, Devin, Ghemawat, Goodfellow, Harp, Irving, Isard, Jia, Jozefowicz, Kaiser, Kudlur, Levenberg, Man\'{e}, Monga, Moore, Murray, Olah, Schuster, Shlens, Steiner, Sutskever, Talwar, Tucker, Vanhoucke, Vasudevan, Vi\'{e}gas, Vinyals, Warden, Wattenberg, Wicke, Yu, and Zheng]{tensorflow2015-whitepaper}
Mart\'{i}n Abadi, Ashish Agarwal, Paul Barham, Eugene Brevdo, Zhifeng Chen, Craig Citro, Greg~S. Corrado, Andy Davis, Jeffrey Dean, Matthieu Devin, Sanjay Ghemawat, Ian Goodfellow, Andrew Harp, Geoffrey Irving, Michael Isard, Yangqing Jia, Rafal Jozefowicz, Lukasz Kaiser, Manjunath Kudlur, Josh Levenberg, Dandelion Man\'{e}, Rajat Monga, Sherry Moore, Derek Murray, Chris Olah, Mike Schuster, Jonathon Shlens, Benoit Steiner, Ilya Sutskever, Kunal Talwar, Paul Tucker, Vincent Vanhoucke, Vijay Vasudevan, Fernanda Vi\'{e}gas, Oriol Vinyals, Pete Warden, Martin Wattenberg, Martin Wicke, Yuan Yu, and Xiaoqiang Zheng.
\newblock {TensorFlow}: Large-scale machine learning on heterogeneous systems, 2015.
\newblock URL \url{https://www.tensorflow.org/}.
\newblock Software available from tensorflow.org.

\end{thebibliography}


\end{document}